\documentclass{article}

\usepackage[final]{nips_2016}

\usepackage[utf8]{inputenc} % allow utf-8 input
\usepackage[T1]{fontenc}    % use 8-bit T1 fonts
\usepackage{hyperref}       % hyperlinks
\usepackage{url}            % simple URL typesetting
\usepackage{booktabs}       % professional-quality tables
\usepackage{amsfonts}       % blackboard math symbols
\usepackage{nicefrac}       % compact symbols for 1/2, etc.
\usepackage{microtype}      % microtypography
\usepackage{amssymb,amsmath,amsthm,amsfonts}

\usepackage{graphicx}
\usepackage{float}
\usepackage{subfigure}

\DeclareMathOperator*{\argmax}{arg\,max}

\title{Nintendo Super Smash Bros. Melee: \\ An "Untouchable" Agent}

\author{
  Ben Parr (bparr), Deepak Dilipkumar (ddilipku), Yuan Liu (yuanl4)\\
Deep Reinforcement Learning and Control (10-703) \\
  Carnegie Mellon University\\
  Pittsburgh, PA 15213
}

\begin{document}

\maketitle

\begin{abstract}
Nintendo's Super Smash Bros. Melee fighting game can be emulated on modern hardware allowing us to inspect internal memory states, such as character positions. We created an AI that avoids being hit by training using these internal memory states and outputting controller button presses. After training on a month's worth of Melee matches, our best agent learned to avoid the toughest AI built into the game for a full minute 74.6\% of the time.
\end{abstract}

\section{Introduction}
Super Smash Bros. Melee is a video game created by Nintendo in 2001 that is still played competitively today. Based on ``current viewership, sponsorship, player base and, most importantly, future growth potential,'' ESPN listed Melee as number six on their 2016 Top 10 Esports Draft (\cite{espn}), making Melee the longest-played game in the Esports Draft. Competitive play is a one player versus another player fighting game where each player chooses one character to play as from the list of 25 playable characters. The goal of the game is to knock the opponent's character off the stage and into the abyss four times before the opponent knocks your character off four times. Each character has a wide array of offensive and defensive abilities. For example, the character Marth has a sword which allows him to attack from a further distance, but at a slower speed. The game was originally created for the Nintendo GameCube console, and so runs at 60 frames per second (fps).

\begin{figure}[H]
  \centering
  \includegraphics[width=.5\textwidth]{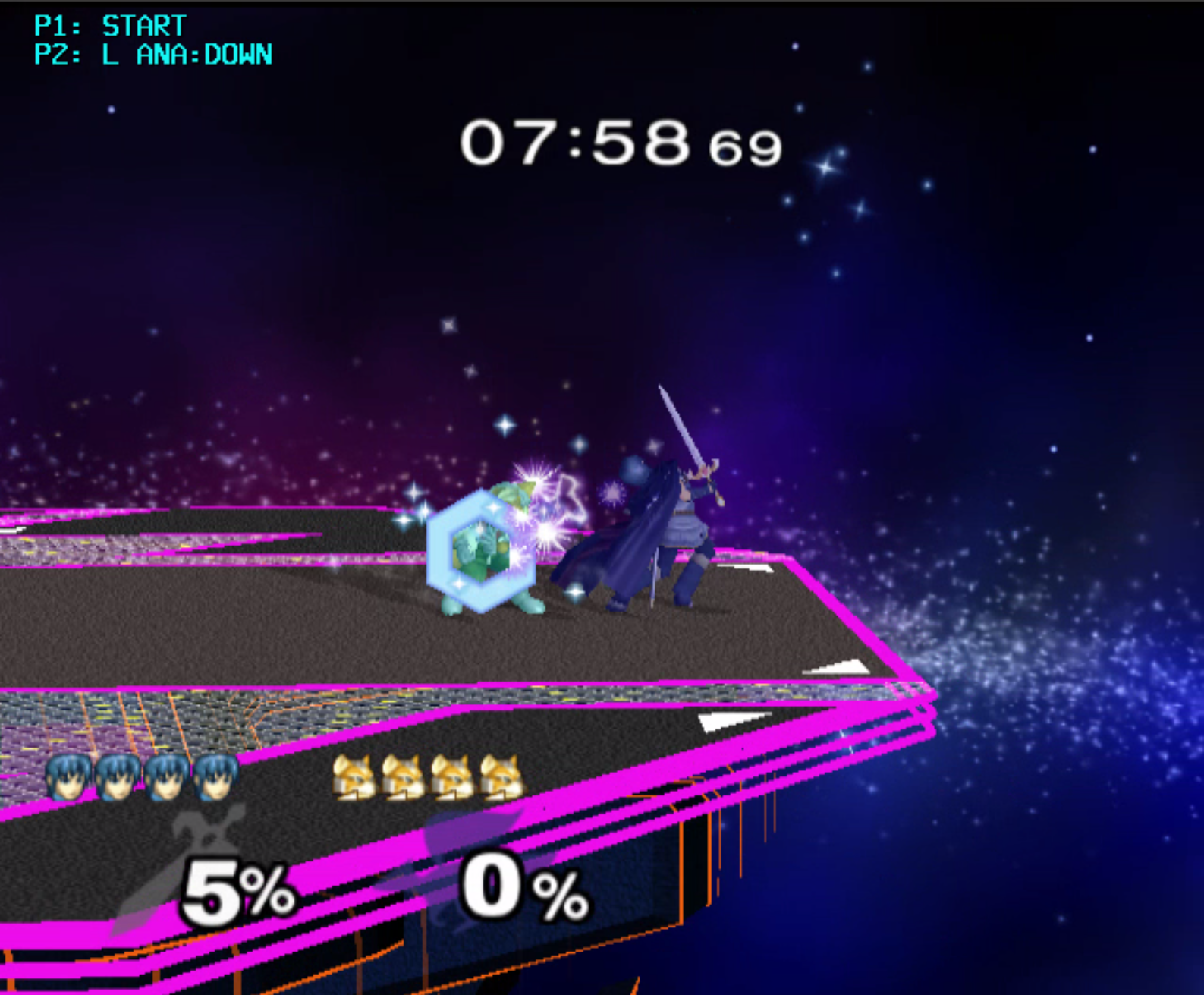}
  \caption{A screenshot of our agent (Fox, left) using a special attack on the sword-wielding opponent (Marth, right) on the entirely flat Final Destination stage.}
\end{figure}

The GameCube is no longer technologically advanced. In fact, Melee can be played on most modern computers using Dolphin software that emulates the GameCube hardware. This allows us to inspect specific memory addresses, such as the memory address containing the position of each character. Fortunately, Vlad Firoiu has written this infrastructure-like code already (\cite{firoiu2017beating}). Firoiu also trained agents for the game using reinforcement learning. For the class project, we decided to train our own independent agents, not based on his reinforcement learning code. Specifically, we trained a Deep Q-Network (DQN) agent, a Double Deep Q-Network (Double DQN) agent, and a Dueling Deep Q-Network (Dueling DQN) agent, as well as a preliminary Asynchronous Advantage Actor-Critic (A3C) agent. Each agent plays as the Fox character and is given the values inside the chosen GameCube memory addresses, and decides which evasive maneuver to take to remain untouched by the opposing Marth character.

Melee has AI built into the game itself, thus allowing a single player option where you play against the built-in computer. The Melee agent has nine levels of difficulty. Unlike our agent, the built-in Melee agent is fixed, and so does not learn from experience. We use the built-in Melee agent to train our own agent. We generated training data in parallel using Google Cloud virtual machines. After training, we evaluated our agents by the number of frames it remained untouched while playing against the toughest AI built into the game.

\section{Related Work}
DeepMind Technologies has successfully learned to play Atari 2600 games using an emulator \cite{mnih2013playing}. DeepMind used a convolutional neural network combined with a Q-learning variant to learn to play games using just the games' pixel output. We plan to build on this success by focusing on a relatively newer game with more complex interactions. Instead of pixels, we look at the game's internal memory, including features such as characters' positions and velocities.

The work most relevant to this project is that of \cite{firoiu2017beating}, which applies Reinforcement Learning to create an agent for Melee. Utilizing an actor-critic model with a deep architecture and self-play, they were able to achieve excellent results with the Captain Falcon character on the Battlefield stage, going on to even beat a number of professional players. Their agent is however known to have difficulties dealing with unusual opponent behavior that has not been seen in training, such as the opponent simply crouching.

There have been more recent papers that build on the DeepMind's earlier success using variations of the regular DQN architecture. One of these is the Double DQN introduced by \cite{double}. This architecture uses two networks playing alternating roles in order to reduce the tendency of the regular DQN to over-estimate action values. Another recent success was the Dueling DQN of \cite{dueling}. This uses two parallel networks that estimate the value function and advantage function, and finally combine these to estimate the Q-value functions.

Another class of algorithms that has gained popularity in the era of Deep Reinforcement Learning is the policy gradient method, such as the actor-critic algorithm. In particular, the Asynchronous Advantage Actor Critic method (A3C) introduced in \cite{a3c} has become one of the standard algorithms for distributed learning in many game-playing applications of reinforcement learning.

\section{Methods}

We implemented four different reinforcement learning algorithms and applied them to Melee. The four algorithms are DQN, Double DQN, Dueling DQN and A3C. Each algorithm interacted with a custom environment, which for example, rewarded taking no damage. Finally, each algorithm trained on Melee gameplay samples generated in parallel using Google Cloud virtual machines.

\subsection{Environment Setup}

We use the Dolphin GameCube emulator to run Melee. The game includes 9 AIs with increasing difficulty, all of which are used during training. Even though our agent predominately trained against the toughest AI, training against all 9 AIs provided our agent a wider amount of combat scenarios. All of our agents play as the Fox character versus the Marth character (controlled by an in-game AI) on Final Destination. This Marth opponent, along with a sporadic frame-skipping issue (see Discussion), make the environment stochastic. 

By looking into the internal game memory, we are able to access information about the current match. These are included in the state that we pass into the algorithm. Specifically, the state includes both the agent's and opposing Marth's current position, velocity, action state, duration of current action state, direction facing, whether charging an attack, whether airborn, shield size, jumps used, hitlag counter, and damage.

The agent has access to five actions to choose from: do nothing, left dodge, right dodge, standing dodge, and Fox's shine (essentially a shield that pushes opponents away). These actions were chosen since our primary goal was to create an agent that could not be hit, for which these defensive actions would be sufficient. The shine attack in particular is tied for the fastest move in the game, and thus should be able to deal with situations that come up which the other moves can't handle.

As for the reward scheme, the agent gets a reward of 1/60 for every frame that it is not moving, and has zero damage. The number 1/60 was chosen to keep the rewards small. We noticed that keeping rewards small in this way helped to stabilize our mean-max Q values. The agent gets a reward of 0 for choosing to dodge or shine, in order to discourage it from spamming these actions. 

The most important aspect of the setup is the terminal flag. As soon as the agent is hit, it receives a terminal flag and the match is reset. This way, it gets a strong negative signal for getting hit, as it can only receive rewards as long as its damage is at 0. This way, the agent only sees frames from the beginning of the game during the start of training. Once it learns to get past these states, it starts to see states from later on. This scheme can be described as a form of curriculum learning, in the sense that the agent has to learn how to fend off attacks later in the game only after it learns to get through the initial frames without being hit.

% Thoughts on calling this curriculum learning? 

\vspace{2ex}

\subsection{Training Algorithms and Specifications}

We implement multiple state-of-the-art Deep Reinforcement Learning techniques to train our agent. The first of these is the DQN introduced in \cite{mnih2013playing}. We approximate the value function $Q_w$ using a deep neural network parametrized by weights $w$ and update the weights as:

$$ w = w + \alpha \left( r + \gamma \max_{a' \in A} Q_w(s',a') - Q_w(s,a) \right) \nabla_w Q_w(s,a)$$

We use target fixing and a replay memory as suggested in the paper. 

Next, we use the Double DQN as described in \cite{double}. This uses the same update rule as for the DQN, but the action selection and value estimation for the target during training are done by two different networks. The online network is used to do action selection, and the target network is used for value estimation. So the update (with $ Q_{w_1} $ as on the online network and $ Q_{w_2} $ as the target network) can be written as :

$$ w_1 = w_1 + \alpha \left( r + \gamma Q_{w_2}(s', \argmax_{a' \in A} Q_{w_1}(s',a')) - Q_{w_1}(s,a) \right) \nabla_{w_1} Q_{w_1}(s,a)$$

This is known to reduce the tendency of the DQN to overestimate its Q-values.

We also use the Dueling DQN as described in \cite{dueling}. The most important aspect of a Dueling DQN is the architecture. It consists of two streams of fully connected layers, one which estimates the state value function, and one which estimates the advantage function. These two streams are combined to get an estimate of the Q-value for a particular (state, action) pair:

$$ Q(s,a) = V(s) + A(s,a) - \frac{1}{|\mathcal{A}|}\sum_{a'} A(s,a')$$ 

After training using the three DQN algorithms, we also experimented to a limited extent with the A3C algorithm introduced in \cite{a3c}. This again involves two ``streams'', both of which are neural networks. One - the actor - estimates the policy directly, as a probability distribution over the actions that can be taken. The other - the critic - is used to estimate state value functions to guide the actor to the correct policy in a principled manner. We have implemented a variant of actor-critic, as described in \cite{a3cnew}, which uses target fixing and a replay memory like in the other methods. This also fits very well with our Google Cloud Parallelization method, which is described in the next section.

Our DQN and Double DQN use two hidden layers, having 128 and 256 units. Dueling and A3C each have two streams. The streams share one hidden layer with 128 units, and then branch out to separate fully connected layers having 512 units. In Dueling, the streams are eventually combined as described above, whereas in A3C, they are used to predict two separate quantities (value function and policy).

All of our experiments use $\gamma = 0.99$, RMSProp for optimization with a learning rate of $0.00025$, batch size of 32, and a linear decay $\epsilon$ greedy policy for exploration, with $\epsilon$ decaying from $1$ to $0.1$ in the first quarter of training. The replay memory size was $1,000,000$, and the target network was updated every $10,000$ training iterations. During training, the agent faces an AI whose level is decided at random. Specifically, we trained against the toughest in-game AI 70\% of the time, with the remaining 30\% distributed across the other AI levels. This was done in order to avoid ``overfitting'' to a particular opponent. See the Discussion section for more information on this.

We use three metrics to compare these methods throughout training.  The first is the reward gained by the agent based on our reward scheme. The second is the average game length that the agent survives for without being hit, in terms of number of frames. The final one is the mean max Q-value, calculated on $1000$ states randomly sampled before training. The third metric gives us an idea of how well the network believes it is performing. The second metric tells us how well the agent is doing in terms of our actual goal - dodging the opponent without getting hit. Finally, links between the first and second metrics will give us a good idea of how well the reward scheme is able to guide our agent to our actual goal. This way, we are able to comprehensively evaluate and compare each of our algorithms.

%Q-learning can be seem as a pseudo stochastic gradient descent step on:
%$$l(w) = E_{s,a,r,s'}\left( r+\gamma \max_{a'\in A} Q_{w^-}(s',a') - Q_{w}(s,a) \right)^2$$

\subsection{Episode Parallelization}

\begin{figure}[H]
\centering
\includegraphics[width=0.6\linewidth]{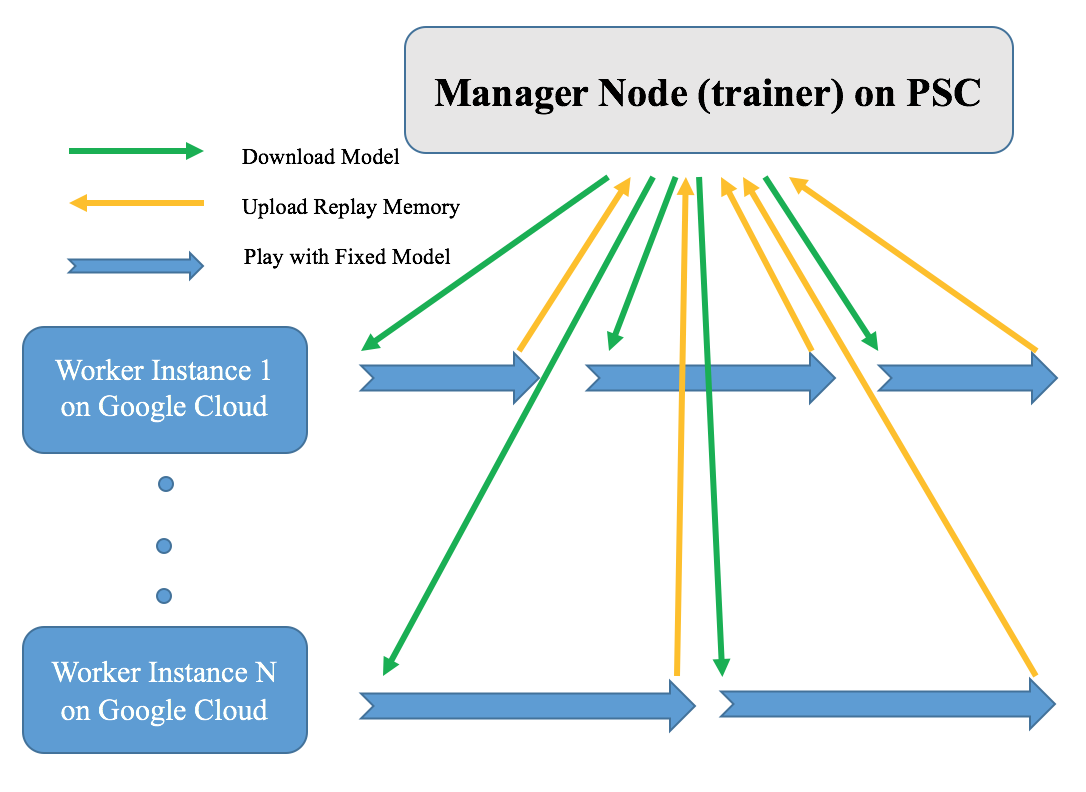}
\caption{Infrastructure outline where a single Pittsburgh Supercomputing Center (PSC) job trains on gameplay generated on N Google Cloud workers}
\end{figure}

A single GPU-enabled manager trains on gameplay data that is generated by 50 workers playing Melee in parallel, allowing each model to train on a full month of gameplay data generated in only 8 hours. First, the manager creates an initial model, and uploads the model to the workers. The workers then enter a loop: generate a few minutes of gameplay samples from the current fixed model, upload the samples to the manager, download the latest model, repeat. Each Melee worker uses a fixed model $Q_{w}$ to generate replay memory tuples ($s,a,r,s'$). This allows the manager to train on a stream of incoming gameplay data generated faster than the manager could generate on its own. Finally, the manager completes the loop by periodically saving an updated model every 15 times the manager receives gameplay samples from its workers.

Fortunately, one of the cheapest available Google Cloud virtual machine, the g1-small machine, is able to run Dolphin emulated Melee. We chose Google Cloud since they provide a free \$300 credit to new users. We were able to run all training for free by only using this \$300 credit.

\section{Experimental Results}

The results of our experiments for DQN, Double DQN and Dueling DQN are shown in Figure 3 and Figure 4. Figure 3 shows rewards and game lengths for all 3 models. First, we see that the results are very erratic, and fluctuate through training. However, all of the models show a clear improvement in performance about 20\% of the way through training.

\begin{figure}[H]
\centering  
\subfigure[DQN Rewards]{\includegraphics[width=0.45\linewidth]{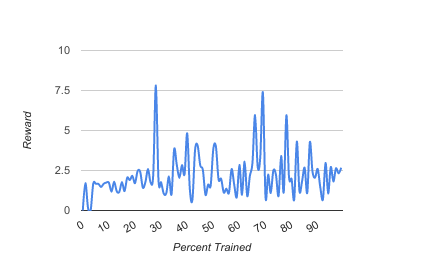}}
\subfigure[DQN Game Lengths]{\includegraphics[width=0.45\linewidth]{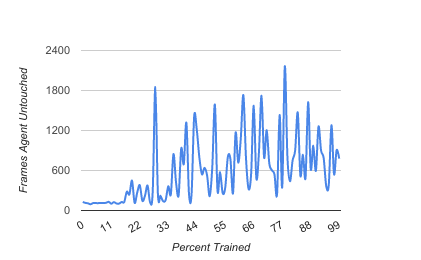}}
\subfigure[Double Rewards]{\includegraphics[width=0.45\linewidth]{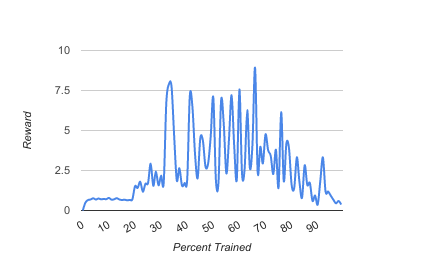}}
\subfigure[Double Game Lengths]{\includegraphics[width=0.45\linewidth]{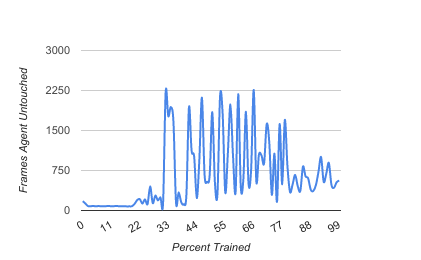}}
\subfigure[Dueling Rewards]{\includegraphics[width=0.45\linewidth]{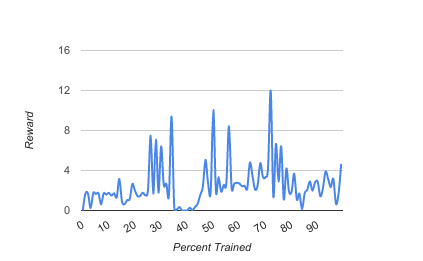}}
\subfigure[Dueling Game Lengths]{\includegraphics[width=0.45\linewidth]{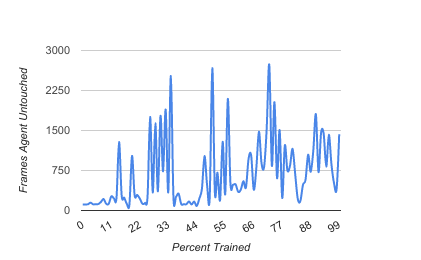}}
\caption{Rewards and Game Lengths for DQN, Double DQN and Dueling DQN}
\end{figure}

The regular DQN does not appear to perform as well as the other two in terms of game lengths. The Double DQN had the most consistently good results, with game lengths remaining high for a significant portion of training. The Dueling DQN however, had the highest peaks, meaning that it produced the best overall agent. Also for context, it is useful to note that at 60 fps, 2250 frames is 37.5 seconds, meaning that each agent is often able to last above 30 seconds without being hit at all for the Double and Dueling networks.

Another useful point to note in these results is the link between the reward and game length. The patterns between these two metrics for each model are quite similar, indicating that our reward scheme is a good proxy to lead the agent towards our goal of dodging the opponent.

Figure 4 shows the Mean Max Q values for these 3 models. The DQN seems to be improving steadily, although for our reward scheme, the value of roughly 2.5 is heavily overestimated. The Double DQN is meant to remedy the DQN's tendency to overestimate its own performance, and we see that it does indeed do that - the mean max Q values for the Double DQN are a lot more reasonable. The graph for Dueling is surprisingly erratic, and almost cyclic in its fluctuations. 

\begin{figure}[H]
\centering
\includegraphics[width=.75\linewidth]{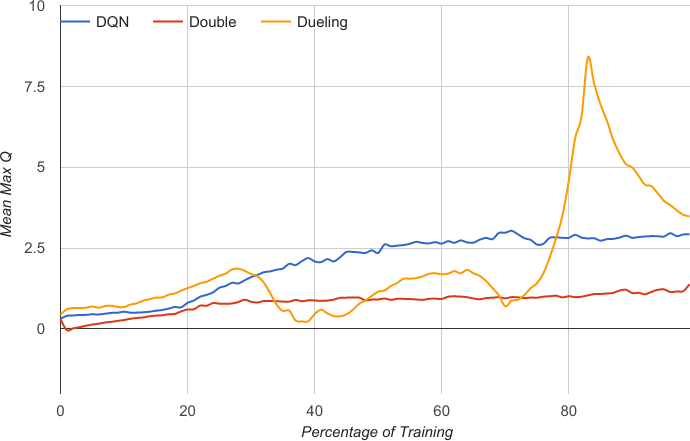}
\caption{Mean Max Q over training.}
\end{figure}

We also include results from our preliminary A3C runs in Figure 5. Note that the third graph shows the estimated value function on the held out states, as opposed to the max Q-value, as it is the value function that the critic evaluates. This is still comparable to the mean max Q-value that we computed for the other methods.

All three graphs indicate the same issue - the agent appears to get stuck early on and is unable to improve any further during training. Watching it playing, we see that it simply chooses to do nothing, and collect the rewards that it can gain in the couple of seconds before it is hit. This is reflected in the game length graph as well, which is stuck at around 120. We believe that this is likely due to training being stuck in a local minimum, since standing still is actually the only way for the agent to increase its reward in the short-term. One possible explanation for this is the need for a more sophisticated exploration strategy. An epsilon greedy policy did not seem to work, and an entropy based exploration scheme (which was used to generate these graphs) also does poorly. Using the right weight for the entropy term in the loss function is also a potential issue. Additionally, using a different optimizer (such as ADAM) might help our agent to break out of this local minimum.

As our experiments with the DQN variants were successful, we did not spend a significant amount time ironing out the issues with the A3C. We realize that it has tremendous potential, especially with the parallelization scheme that we already use, and so we hope to solve these problems and get A3C working well in the future.

Videos of our agent playing can be found at \url{https://goo.gl/x67ioE}. These are all minute-long matches between our agent (Fox) and the highest level AI in the game (Marth), in which our agent completely avoids being hit. Our best agent, trained using the Dueling DQN, is able to dodge a level 9 AI for a minute or longer, 74.6\% of the time.

\begin{figure}[H]
\centering  
\subfigure[Rewards]{\includegraphics[width=0.47\linewidth]{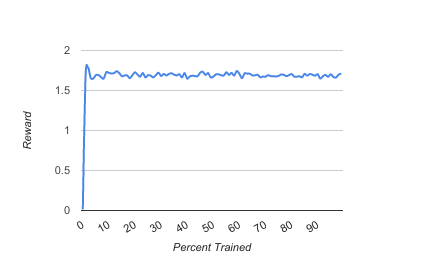}}
\subfigure[Game Lengths]{\includegraphics[width=0.47\linewidth]{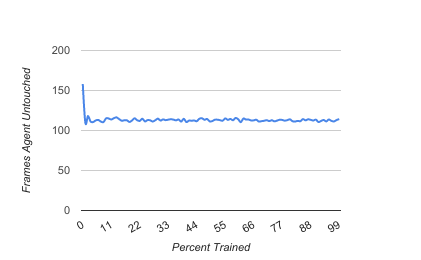}}
\subfigure[Value Function]{\includegraphics[width=0.47\linewidth]{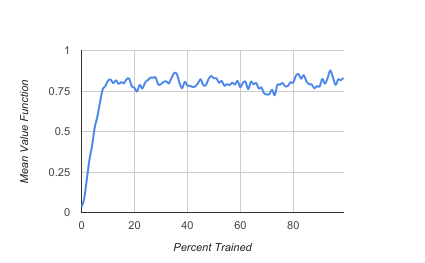}}

\caption{A3C results}
\end{figure}

\section{Discussion}

The major limitation of our project was the opposing player. Of course, we would have preferred training against human players who play Melee professionally. Based on our resources, we instead trained against the AI players built into the game itself. So the agent ``overfits'' the toughest in-game AI's strategy, which we were able to alleviate to some extent by randomizing the CPU level during training.

One solved issue was that our initial DQN models learned to do nothing, and so, would be hit by the opponent Marth almost immediately. We solved this issue by not using a ReLU activation on the final layer of our deep networks. Interestingly, a ReLU on the final layer seemed to completely block learning in Melee, where initially zero Q-values never change during training. One open issue is the erratic Dueling DQN Q-values, which spike above 8.0 after 80\% training. Reducing the target update frequency could help make training more stable and produce a smoother mean max Q.

Finally, we observed sporadic frame skipping issues when emulating Melee, where our agent would hit a button but Melee would ignore it. This makes Melee harder for our agent. Future work could look into technology used by Tool-Assisted Speedrun (TAS) to remove the frame skipping issue, and thus make the opponent player the only stochastic part of the environment.

\section{Conclusion}
DQN performs worse than Double DQN and Dueling DQN. Double DQN showed consistent performance during training, in terms of game lengths and rewards, and also has reasonable Q-values. However, the Dueling DQN had the highest peaks on the game lengths graph, meaning that it produced the best agents. After training on one month of gameplay data with the Dueling DQN, our best agent is able to dodge the highest level AI in the game for at least a full minute 74.6\% of the time.

Our code can be found at \url{https://github.com/bparr/melee-ai} and videos of our agent playing can be found at \url{https://goo.gl/x67ioE}.

\section{Acknowledgements}
Thank you to Vlad Firoiu, without whom this class project would not be possible (\cite{firoiu2017beating}). All the best at Google DeepMind.

\bibliographystyle{apalike}
\nocite{*} % Include all non-cited references.
\bibliography{bib}

\end{document}